\documentclass{article}
\usepackage{spconf,amsmath,graphicx}
\usepackage{amssymb}
\usepackage{latexsym}
\usepackage{exscale}
\usepackage{fontenc}
\usepackage{graphicx}
\usepackage{epsfig}
\usepackage{nicefrac}
\usepackage{mathptmx}
\usepackage{hyperref}

\title{Online Adaptive Statistical Compressed Sensing \\ of Gaussian Mixture Models}

\twoauthors
 {Julio Duarte-Carvajalino, Guillermo Sapiro, Guoshen Yu}
    {University of Minnesota
  }
  {Lawrence Carin}
  {Duke University
}

\begin{document}

\renewcommand{\baselinestretch}{1.1}
\newcommand{\nin}{\in\!\!\!/}
\newcommand{\rb}{\rangle}
\newcommand{\lb}{\langle}
\newcommand{\R}{{\bf R}}
\newcommand{\thTh}{{\theta \in \Theta}}
\newcommand{\Dip}{{D}}
\newcommand{\cG}{{\cal G}}
\newcommand{\cQ}{{\cal Q}}
\newcommand{\curvelet}{{c}}
\newcommand{\orient}{{\theta}}
\newcommand{\ERC}{{\rm ERC}}
\newcommand{\DU}{{{\cal D}_U}}
\newcommand{\ipidx}{{p}}
\newcommand{\ipatom}{{g}}
\newcommand{\Ga}{{\Gamma}}
\newcommand{\pGa}{{p \in \Gamma}}
\newcommand{\lu}{{\bf l}^1}
\newcommand{\Ld}{{\bf L}^2}
\newcommand{\lz}{{\bf l}^0}
\newcommand{\ld}{{\bf l}^2}
\newcommand{\V}{{\bf V}}
\newcommand{\cL}{{\cal L}}
\newcommand{\W}{{\bf W}}
\newcommand{\La}{{\Lambda}}
\newcommand{\tLa}{{\tilde \Lambda}}
\newcommand{\pLa}{{p \in \Lambda}}
\newcommand{\qO}{{q \in \Omega}}
\newcommand{\ptLa}{{p \in \tilde \Lambda}}
\newcommand{\supp}{{\Lambda}}
\newcommand{\om}{{\omega}}
\newcommand{\mq}{{m,q}}
\newcommand{\jn}{{j,n}}
\newcommand{\opt}{{\tilde}}
\newcommand{\IPop}{{U}}
\newcommand{\CC}{{\bf C}}
\newcommand{\Z}{{\bf Z}}
\newcommand{\Reg}{\Lambda}
\newtheorem{conjecture}{Conjecture}
\newtheorem{definition}{Definition}
\newtheorem{theorem}{Theorem}
\newtheorem{lemma}{Lemma}
\newtheorem{corollary}{Corollary}
\newtheorem{prop}{Proposition}
\providecommand{\argmin}{\mathop{\textup{argmin}}}
\def\truelabel{\label}
\newcommand{\Span}{\mathop{\textup{span}}}
\newcommand{\pen}{\text{pen}}
\newcommand{\ud}{\textup{d}}
\newcommand{\D}{\mathcal{D}}
\newcommand{\Mg}{\mathcal{M}_{\gamma}}
\newcommand{\Tde}{T^{\frac{2\alpha}{\alpha+1}}}
\newcommand{\B}{\mathcal{B}}
\newcommand{\wor}{k}
\newcommand{\Cal}{\mathbf{C}^{\alpha}}
\newcommand{\eqdef}{\overset{.def}{=}}
\newcommand{\tga}{\tilde g}
\newcommand{\ldeuxj}{l^2_j}
\newcommand{\ga}{g}
\newcommand{\nogeom}{\Xi}
\newcommand{\tS}{\tilde S}
\newcommand{\tb}{\tilde b}
\newcommand{\ldeux}{l^2}
\newcommand {\ImU} {{\bf ImU}}
\newcommand{\Proba}{\mathbb{P}}
\newcommand{\C} {{\bf C}}
\newcommand{\gammageom}{\upsilon}
\newcommand{\Ch}[1]{{\bf Ch: #1}}

\newcommand{\ba}{\mathbf{a}}
\newcommand{\bb}{\mathbf{b}}
\newcommand{\bbf}{\mathbf{f}}
\newcommand{\bx}{\mathbf{x}}
\newcommand{\by}{\mathbf{y}}
\newcommand{\bw}{\mathbf{w}}
\newcommand{\bz}{\mathbf{z}}

\newcommand{\bzero}{\mathbf{0}}

\newcommand{\bA}{\mathbf{A}}
\newcommand{\bB}{\mathbf{B}}
\newcommand{\bC}{\mathbf{C}}
\newcommand{\bD}{\mathbf{D}}
\newcommand{\bI}{\mathbf{I}}
\newcommand{\bM}{\mathbf{M}}
\newcommand{\bR}{\mathbf{R}}
\newcommand{\bS}{\mathbf{S}}
\newcommand{\bT}{\mathbf{T}}
\newcommand{\bU}{\mathbf{U}}
\newcommand{\bW}{\mathbf{W}}

\newcommand{\sample}{\Phi}
\newcommand{\dict}{\Psi}
\newcommand{\expect}{E}
\newcommand{\nsp}{\mathrm{Null}}

\maketitle

\begin{abstract}
A framework of online adaptive statistical compressed sensing is introduced
for signals following a mixture model. The scheme first uses non-adaptive measurements, 
from which an online decoding scheme estimates the model selection. As soon as a candidate model has 
been selected, an optimal sensing scheme for the selected model continues to apply. The final signal reconstruction 
is calculated from the ensemble of both the non-adaptive and the adaptive measurements. For signals generated from a Gaussian
mixture model, the online adaptive sensing algorithm is given and its performance is analyzed. On both synthetic and 
real image data, the proposed adaptive scheme considerably reduces the average
reconstruction error with respect to standard statistical compressed sensing
that uses fully random measurements, at a marginally increased computational complexity. 
\end{abstract}

 
\section{Introduction}
\vspace{-2ex}
Compressed sensing (CS) aims at achieving accurate signal reconstruction while sampling signals at a low sampling rate, typically far smaller than that of Nyquist. Let $\bx \in \mathbb{R}^N$ be a signal of interest, $\Phi \in \mathbb{R}^{M \times N}$ a \textit{non-adaptive} sensing matrix (\textit{encoder}), consisting of $M \ll N$ measurements, $\by = \Phi \bx \in \mathbb{R}^M$ a measured signal, and $\Delta$ a \textit{decoder} used to reconstruct $\bx$ from $\Phi \bx$. CS develops encoder-decoder pairs $(\Phi, \Delta)$ such that a small reconstruction error $\|\bx - \Delta(\Phi \bx)\|_X$, where $ \|\cdot\|_X$ is a norm, can be achieved. 

Assuming a sparse signal model, i.e., the signal can be accurately represented in a dictionary with a few non-zero coefficients, the CS theory has shown that using random sensing matrices such as Gaussian or Bernoulli matrix $\Phi$ with $M=\mathcal{O}(k \log(N/k))$ measurements, and an $l_1$ minimization or a greedy matching pursuit decoder $\Delta$ promoting sparsity, with high probability accurate signal reconstruction is possible: The obtained approximation error $\|\bx - \Delta(\Phi \bx)\|_X$ is tightly upper bounded by a constant times the best $k$-term approximation error in the sparse representation~\cite{cohen2009compressed}. 

While conventional CS deals with one signal at a time, statistical compressed sensing (SCS) aims at efficiently sampling a collection of signals and having accurate reconstruction on average. Assuming that the signals $\bx$ follow a distribution with probability density function (pdf) $f(\bx)$, SCS designs encoder-decoder pairs $(\sample, \Delta)$ so that the \textit{average} error $
 \expect \|\bx - \Delta(\Phi \bx) \|_X = \int  \|\bx - \Delta(\Phi \bx) \|_X f(\bx) d\bx \nonumber
$ is small~\cite{yu2011SCS}. 

For signals following a Gaussian distribution, it has been shown that with \textit{any} sensing matrix of $M=k$ measurements and the maximum a posteriori (MAP) linear decoder, SCS leads to a mean squared error (MSE)  $\expect \|\bx - \Delta(\Phi \bx) \|^2$ upper bounded by a constant times the minimum MSE obtained with the $k$-term linear approximation in the principal direction analysis (PCA) basis that is optimal for Gaussian signals~\cite{yu2011SCS}. In particular, the error bound is tight when Gaussian or Bernoulli random sensing matrix is used~\cite{yu2011SCS}. For signals generated from a Gaussian mixture model (GMM), i.e., there exist multiple Gaussian distributions and each signal is generated from one of them with an unknown index, GMMs giving more precise description of most real signals than single Gaussian models, a piecewise linear decoder that calculates the signal reconstruction from each of the Gaussian models, and then selects the best one, has been introduced~\cite{yu2011SCS, yu2010PLE}. Additional theoretical results on the Gaussian model selection accuracy and overall reconstruction have been shown in~\cite{chen2010compressive}. 

SCS of GMM applies \textit{non-adaptive} random sensing matrices because for the signal being sensed, the Gaussian model from which the signal is generated is a priori unknown. If it were known, one would then prefer sensing along the principal directions in the appropriate Gaussian, which leads to the minimum MSE.\footnote{This optimal MSE sensing for Gaussians is easy to prove, see next Section, while the optimal sensing for other distributions has been recently elegantly developed in~\cite{Carson2011}. The  strategy here introduced can then be extended to mixtures beyond GMMs.} More generally speaking, assume that the signals are generated from a mixture model, and an optimal sensing scheme is associated with each of the underlying models (e.g., following~\cite{Carson2011}). If for the current signal, its model were known before sensing, the optimal sensing scheme in that distribution would be preferred rather than using non-adaptive measurements. 

This paper follows this line of thoughts and introduces an online adaptive sensing framework for signals generated from a mixture model. The scheme imbeds an online model selection and a switch from \textit{non-adaptive} to \textit{adaptive} sensing. To sense a signal, non-adaptive measurements are first used, from which an online decoding scheme calculates the model selection. As soon as a model has been selected, the optimal sensing scheme of the selected model then continues to apply. The final signal reconstruction is calculated from the ensemble of both the non-adaptive and the adaptive measurements. 

As an important example, this online adaptive sensing is here illustrated for signals following a GMM. Not only GMMs have been shown to lead to results in the ballpark of the state-of-the-art in various inverse problems for different types of real data~\cite{leger2010Matrix, yu2010PLE}, theoretical results on statistical compressed sensing of GMM have also been recently given~\cite{chen2010compressive, yu2011SCS}.

Section~\ref{sec:SCS:GMM} recalls the main results of SCS of GMM~\cite{yu2011SCS} based on which the online SCS of GMM will be developed. An algorithm for the online adaptive SCS of GMM is then given in Section~\ref{sec:online:SCS}, and its performance is analyzed and compared against standard SCS using fully random sensing. In Section~\ref{sec:experiments} the  proposed online adaptive SCS is applied in real image data sensing, leading to considerably improved results with respect to standard SCS, at a marginally increased computational complexity. Concluding remarks and future work are discussed in Section~\ref{sec:conclusion}.

\section{Statistical Compressed Sensing}
\label{sec:SCS:GMM}
\vspace{-1ex}
\subsection{Sensing of Gaussian Models}
\vspace{-1ex}
\subsubsection{Optimal Principal Direction Sensing}
\vspace{-1ex}
Signals $\bx \in \mathbb{R}^N$ are assumed to follow a Gaussian distribution $\mathcal{N}(\mu, \Sigma)$, where $\mu$ and $\Sigma$ are respectively its mean and covariance.  Without loss of generality, the Gaussian mean is assumed zero, ${\mu} = \mathbf{0}$, as one can always center the signal with respect to the mean.  Principal Component Analysis gives the orthonormal PCA basis $\bB$ that diagonalizes the covariance matrix $\Sigma = \bB \bS\bB^T$,
where $\bS = \mathrm{diag}(\lambda_{1}, \ldots, \lambda_{N})$ is a diagonal matrix whose diagonal elements $\lambda_1 \geq \lambda_2 \geq \ldots \geq \lambda_N$ are the sorted eigenvalues~\cite{mallat2008wts}. It is well known that for Gaussian signals a linear approximation in the PCA basis minimizes the mean squared error (MSE).
Putting this in the signal sensing context, a sensing matrix
\vspace{-1ex}
\begin{equation}
\label{eqn:PCA:sensing}
\Phi  = [\bb_1, \ldots, \bb_M]^T\in \mathbb{R}^{M \times N},\vspace{-1ex}
\end{equation}
where $\bb_n$ is the $n$-th principal direction of the Gaussian, i.e., the $n$-th column in $\bB$, and a linear decoder 
\vspace{-1ex}
\begin{equation}
\label{eqn:PCA:reconstruction}
\Delta = [\bb_1, \ldots, \bb_M]  \in \mathbb{R}^{N \times M}, \vspace{-1ex}
\end{equation}
 minimize the MSE amongst all sensing matrices 
$\Phi \in \mathbb{R}^{M \times N}$ and any decoder $\Delta$:
\vspace{-2ex}
\begin{equation}
\label{eqn:linear:approximation}
\hspace{-1ex}\small {\sigma^2_M \triangleq \min_{\Phi \in \mathbb{R}^{M \times N}, \Delta} E[\|\bx - \Delta (\Phi {\bx})\|^2]  
= E[\|\bx - \sum_{n=1}^M \langle \bx, \bb_n \rangle \bb_n \|^2] = \sum_{n=M+1}^N \lambda_n,} \nonumber
\vspace{-2ex}
\end{equation}
where $\sigma^2_M$ denotes the minimum MSE. 
\vspace{-1ex}

\subsubsection{Statistical Compressed Sensing}
\vspace{-1ex}
For Gaussian signals, it has been shown that \textit{any} sensing matrix $\Phi \in \mathbb{R}^{M \times N}$ and the maximum a posteriori (MAP) linear decoder $\Delta = \Sigma \Phi^T (\Phi \Sigma \Phi^T)^{-1}$ lead to an MSE upper bounded by a constant times the minimum MSE~\cite{yu2011SCS}: 
\vspace{-1ex}
\begin{theorem}
\label{theo:SCS:with:RIP:expect}
Assume $\bx \sim \mathcal{N}(\bzero, \Sigma)$. Let $\Phi$ be an $M \times N$ sensing matrix and $\Delta= \Sigma \Phi^T (\Phi \Sigma \Phi^T)^{-1}$ the optimal and linear decoder. Then
\begin{equation}
\label{eqn:instance:optimality:MSE:firstk}
E [\|\bx - \Delta (\Phi \bx)\|_2^2] \leq C_0 \sigma^2_M,
\end{equation}
where the constant $C_0$ is defined in~\cite{yu2011SCS}. 
\end{theorem}
The bound constant  $C_0$ in Theorem~\ref{theo:SCS:with:RIP:expect} can be obtained via Monte Carlo simulations. For Gaussian and Bernoulli matrices,  a small $C_0\approx 4.5$ has been shown, i.e., the error bound is tight~\cite{yu2011SCS}. 
\vspace{-1ex}
\subsection{Sensing of Gaussian Mixture Models}
\label{subsec:sensing:GMM}
\vspace{-1ex}
A single Gaussian distribution is often too simplistic for modeling real signals. Assuming multiple Gaussian distributions $\{\mathcal{N} (\bzero, \Sigma_j)\}_{1 \leq j \leq J}$
and that each signal follows one of them with an unknown index, Gaussian mixture models (GMMs) provide more precise signal descriptions. 

As the Gaussian indices of the signals are unknown, the optimal principal sensing~\eqref{eqn:PCA:sensing},~\eqref{eqn:PCA:reconstruction} is impracticable. SCS applies instead non-adaptive random matrices for signal sensing and a piecewise linear decoder for reconstruction~\cite{yu2011SCS}. The piecewise linear decoder  first calculates
the linear MAP decoder using each of the Gaussian models, \vspace{-2ex}
\begin{equation}
\label{eqn:MAP:Sigma:K} 
\tilde{\bx}_j \triangleq \Delta_j (\Phi \bx) = \Sigma_j \Phi^T (\Phi \Sigma_j \Phi^T)^{-1} (\Phi \bx),~~~\forall  1 \leq j \leq J,
\vspace{-1ex}
\end{equation}
and then selects the best model $\tilde{j}$ that maximizes the log  a-posteriori probability  among all the models~\cite{yu2010PLE}
\vspace{-1ex}
\begin{equation}
\label{eqn:GMM:model:selection} 
\tilde{j} = \arg \max_{1 \leq j \leq J} -\frac{1}{2} \left(\log |\Sigma_{j}| + \tilde{\bx}_j^T \Sigma_{j}^{-1} \tilde{\bx}_j\right),
\vspace{-1ex}
\end{equation}
whose corresponding decoder $\Delta_{\tilde{j}}$ gives the final signal reconstruction: 
\vspace{-1ex}
\begin{equation}
\label{eqn:best:decoder} 
\Delta(\Phi \bx)  = \Delta_{\tilde{j}} (\Phi \bx). 
\vspace{-1ex}
\end{equation}

The accuracy of the Gaussian model selection~\eqref{eqn:GMM:model:selection} and of the signal reconstruction given by the piecewise linear decoder has been shown influenced by a number of factors, including the geometry
of the Gaussian distributions in the GMM, the signal dimension, and the number of sensing measurements~\cite{yu2011SCS}. More accurate model selection and smaller reconstruction error is obtained as the Gaussians distributions are more ``orthogonal'' one another, as each of the Gaussians is more anisotropic, as the signals are in a higher dimension given that the energy of the signals are concentrated in the first few dimensions, and as the number of sensing measurements increases. Additional theoretical results on Gaussian model selection have been given in~\cite{chen2010compressive}.

\vspace{-1ex}
\section{Online Adaptive Statistical Compressed Sensing}
\label{sec:online:SCS}

\vspace{-1ex}

SCS of GMM applies \textit{non-adaptive} random sensing matrices because for the signal to be sensed, the Gaussian model from which the signal is generated is a priori unknown. If it were known, one would then prefer sensing along the principal directions in the appropriate Gaussian, which leads to the minimum MSE. 

The online adaptive SCS improves the accuracy of SCS by first selecting online the Gaussian model, and then adapts the measurements as a function of the model selection. It starts by performing non-adaptive random measurements, based on which the piecewise linear decoder estimates online the Gaussian model for the signal being sensed. As soon as the Gaussian model is selected, for the rest of the measurements it switches to the principal direction sensing in the selected Gaussian. As long as the online model selection is correct, the adaptive sensing along the principal directions in the appropriate Gaussian leads to a smaller MSE than applying fully random sensing.  
\vspace{-5ex}

\subsection{Algorithm}
\vspace{-1ex}
Assume that $M \leq N$ measurements are dedicated to sensing the signal. The online SCS algorithm proceeds as follows. 
\vspace{-1ex}
\begin{enumerate}
\item \textbf{Random sensing}. Sense the signal with a random matrix $\Phi_K^R \in \mathbb{R}^{K \times N}$ of $K \leq M$ measurements. 
\item \textbf{Online decoding and model selection}. Decode online the signal from $\Phi_K^R \bx$ using the piecewise linear decoder~\eqref{eqn:MAP:Sigma:K} and~\eqref{eqn:GMM:model:selection}:
\vspace{-1ex}
\begin{equation}
\label{eqn:MAP:Sigma:K:online} 
\hspace{-2ex}\tilde{\bx}_j^R \triangleq \Delta_j (\Phi^R_K \bx) = \Sigma_j \Phi^T (\Phi^R_K \Sigma_j (\Phi^R_K)^T)^{-1} (\Phi^R_K \bx),~~~\forall  1 \leq j \leq J,
\vspace{-1ex}
\end{equation}
\vspace{-1ex}
\begin{equation}
\label{eqn:GMM:model:selection:online} 
\hat{j} = \arg \max_{1 \leq j \leq J} -\frac{1}{2} \left(\log |\Sigma_{j}| + (\tilde{\bx}_j^R)^T \Sigma_{j}^{-1} \tilde{\bx}_j^R\right).
\vspace{-2ex}
\end{equation}
\item \textbf{Optimal sensing}. Sense the signal with $\Phi^{\hat{j}}_{M-K} = [\bb^{\hat{j}}_1, \ldots, \bb^{\hat{j}}_{M-K}]^T \in \mathbb{R}^{(M-K) \times N}$, i.e., the first $M-K$ first principal direction vectors in the $\hat{j} $-th Gaussian selected online in~\eqref{eqn:GMM:model:selection:online}.  
\item \textbf{Decoding}. Write $\Phi \bx = [(\Phi_K^R \bx)^T, (\Phi^{\hat{j}}_{M-K} \bx)^T]^T \in \mathbb{R}^{M \times 1}$ the concatenation of the signal measurements sensed in steps 1 and 3. Decode the signal from $\Phi \bx$ with the piecewise linear decoder~\eqref{eqn:MAP:Sigma:K},~\eqref{eqn:GMM:model:selection},~and~\eqref{eqn:best:decoder}.
\end{enumerate} 
\vspace{-1ex}

Contrary to the conventional CS and SCS that apply linear sensing, the sensing of the online adaptive SCS is \textit{nonlinear}, as the principal direction sensing matrix $\Phi_{M-K}$ in Step 3 depends on the Gaussian model selection estimated from the random measurements sensed in Step 1. 

The online adaptive sensing algorithm marginally increases the computational complexity with respect to standard SCS using fully random measurements. The sensing complexity is the same, but the online SCS has an additional online decoding step.  The complexity of decoding~\eqref{eqn:MAP:Sigma:K} is dominated by the $M \times M$ matrix inversion, which requires $M^3/3$ floating-point operations (flops)~\cite{yu2010PLE}. With a GMM comprised of $J$ Gaussian distributions, the online and the final decoding steps are respectively calculated in $JK^3/3$ and $JM^3/3$ flops. As $K \leq M$, the additional online decoding brings a marginal increase in computational complexity. 

Adjusting the number $K$ of random measurements in the online SCS trades off between the online Gaussian model selection accuracy in Step 2 and the signal reconstruction error in Step 4. The larger the $K$, the more random measurements are dedicated, and more accurate the online Gaussian model selection is in consequence (see~\cite{chen2010compressive} for the exact bounds).  Given the correct online Gaussian model selection, a smaller $K$ leaves a bigger number $M-K$ measurements along the principal directions of the appropriate Gaussian, which reduces the signal reconstruction error. 

\vspace{-1ex}
\subsection{Performance Analysis}
\label{subsec:performance:analysis}
\vspace{-1ex}
To better understand the performance of the online adaptive SCS, let us analyze a GMM comprised of two Gaussian distributions $\mathcal{N}(\bzero, \Sigma_1)$ and $\mathcal{N}(\bzero, \Sigma_2)$. Assume without loss of generality that the signals follow the first Gaussian distribution $\bx \sim \mathcal{N}(\bzero, \Sigma_1)$. 
The MSE of the online SCS can be written as 
\vspace{-1ex}
\begin{equation}
\label{eqn:online:SCS:MSE}
\small{\hspace{-3ex}E\|\bx - \Delta (\Phi \bx)\|^2 = \sum_{\hat{i}=1}^2 \sum_{\tilde{i}=1}^2 \int_{\hat{j}=\hat{i}~\textrm{and}~\tilde{j}=\tilde{i}}  \|\bx - \Delta_{\tilde{j},\hat{j}} (\Phi_{\hat{j}}^{(K)} \bx)\|^2 f_1(\bx) d \bx}
\vspace{-1ex}
\end{equation}
where $f_1(\bx) = \frac{1}{(2\pi)^{N/2} |\Sigma_1|^{1/2}} \exp\left({-\frac{1}{2} \bx^T \Sigma_1^{-1} \bx}\right)$,  
$\hat{j}$ and $\tilde{j}$ index respectively the Gaussian model selected online and at the final signal reconstruction,
$\Phi_{\hat{j}}^{(K)} = 
 [(\Phi_K^R)^T, (\Phi^{\hat{j}}_{M-K})^T]^T$
is the concatenation of the random sensing matrix $\Phi_K^R$ of $K$ measurements and the principal direction
sensing matrix $\Phi^{\hat{j}}_{M-K} = [\bb^{\hat{j}}_1, \ldots, \bb^{\hat{j}}_{M-K}]^T$ of $M-K$ measurements in the Gaussian selected online,
and $\Delta_{\tilde{j}, \hat{j}} = \Sigma_{\tilde{j}} (\Phi_{\hat{j}}^{(K)})^T (\Phi_{\hat{j}}^{(K)}  \Sigma_{\tilde{j}} (\Phi_{\hat{j}}^{(K)} )^T)^{-1}$. 
\eqref{eqn:online:SCS:MSE} includes 4 components: 
\vspace{-1ex}
\begin{enumerate}
\item $\hat{j}=1~\textrm{and}~\tilde{j}=1$: Both the online decoding in Step 2 and the final
decoding in Step 4 correctly select the Gaussian model for the signal.  
\item $\hat{j}=1~\textrm{and}~\tilde{j}=2$: The online decoding correctly selects
the Gaussian model, whereas the final decoding incorrectly selects the Gaussian model. 
\item $\hat{j}=2~\textrm{and}~\tilde{j}=1$: The online decoding incorrectly selects
the Gaussian model, whereas the final decoding correctly selects the Gaussian model. 
\item $\hat{j}=2~\textrm{and}~\tilde{j}=2$: Both the online decoding in Step 2 and the final
decoding in Step 4 incorrectly select the Gaussian model for the signal.  
\end{enumerate}
\vspace{-1ex}

To further understand the behavior of the four error components, Monte Carlo simulation is performed
to check them on synthetic data. The data set up follows that in~\cite{yu2011SCS}, emulating standard behavior of image patches: 
the signals are of dimension $N=64$; the eigenvalues of the Gaussians
follow a power decay law $\lambda_m = m^{-\alpha},~1 \leq m \leq N,$ with a typical value $\alpha=2$; the two 
Gaussians are ``orthogonal'' one another, i.e., $\bB_1^T \bB_2 = \bI_{lr}$, where $\bB_1$ and $\bB_2$ are
the PCAs of the two Gaussians, and $\bI_{lr}$ is the left-right flipped identity matrix. The sensing matrix $\Phi$
contains $M=16$ measurements (sampling rate 1/4), and the number $K$ of random measurements varies from $1$ to $M$. The Gaussian model
selection is more accurate as the two Gaussians are orthogonal (see also~\cite{chen2010compressive}). On the other hand, when the online model selection is
erroneous, the resulting principal direction sensing in the wrong Gaussian is the farthest possible from optimal. 

\begin{figure}[htbp]
\vspace{-11ex}
\begin{center}
\begin{tabular}{c}
\includegraphics[width=5cm]{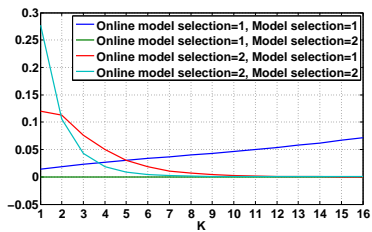}  
\end{tabular}
\end{center}
\vspace{-15ex}
\caption{\small{The four error components of the online SCS as a function of $K$, the number of the first-step random measurements.}} \label{fig:MSE:4parts:OnlineSCS}
\vspace{0ex}
\end{figure}

Figure~\ref{fig:MSE:4parts:OnlineSCS} plots the four error components (normalized by the signal energy) as a function 
of the number $K$ of random measurements in the online SCS. The first component increases as $K$ increases: 
As the online model selection is correctly calculated from the $K$ random measurements, $M-K$ principal direction
vectors of that Gaussian are then used to sense the signal; a larger $K$ leads to a smaller number of $M-K$
optimal principal direction sensing measurements, and thus to a larger error. The second component is constantly
zero: If the online decoding correctly calculates the model selection from the $K$ random measurements, after adding 
$M-K$ measurements along the principal directions in the appropriate Gaussian, the model selection in
the final decoding never goes wrong. The third and the fourth components decrease as $K$ increases: 
The incorrect model selection obtained by the online decoding from 
 the first $K$ random measurements leads to $M-K$ principal direction measurements in the wrong Gaussian; in our example, the 
 two Gaussians have the opposite eigenvalue order, and these $M-K$ principal direction measurements in the wrong Gaussian 
 are therefore the worst possible; a larger $K$ reduces the number $M-K$ of the principal direction measurements in the wrong Gaussian, 
 and reduces in consequence the error, since having more random measurements is better than having more of the wrong measurements. 
 
\begin{figure}[htbp]
\vspace{-13ex}
\begin{center}
\begin{tabular}{c}
\includegraphics[width=5cm]{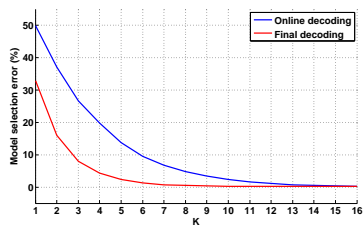}  
\end{tabular}
\end{center}
\vspace{-15ex}
\caption{\small{The Gaussian model selection error (\%) of the online and final coding as a function of  $K$.}}
\label{fig:ModelSelectionError:OnlineSCS}
\vspace{-3ex}
\end{figure}

 Figure~\ref{fig:ModelSelectionError:OnlineSCS} plots the Gaussian model selection errors of the online and final coding
  as a function of $K$. Both errors decrease as $K$ increases. The online model selection
  error is constantly larger than that of the final model selection, the two converging as $K$ goes to $M$. 
   
\begin{figure}[htbp]
\vspace{-14ex}
\begin{center}
\begin{tabular}{c}
\includegraphics[width=5cm]{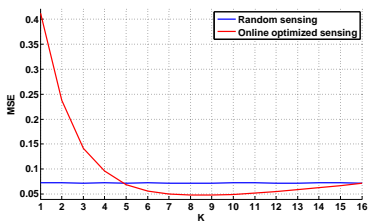}  
\end{tabular}
\end{center}
\vspace{-15ex}
\caption{\small{The MSE of online adaptive SCS as a function of  $K$, in comparison to that of SCS using fully random sensing matrices.}} 
\label{fig:MSE:OnlineSCS}
\vspace{-1ex}
\end{figure}

The sum of the four online SCS error components illustrated in Figure~\ref{fig:MSE:4parts:OnlineSCS} gives the MSE of the online SCS, plotted in Figure~\ref{fig:MSE:OnlineSCS} (red) as a function of $K$. The curve presents a U-shape. When $K$ is small, the
online model selection is inaccurate, the principal direction sensing is thus likely in the wrong Gaussian, which results in a large MSE. As $K$ increases, 
the MSE first decreases and then increases, trading off between the online model selection accuracy (the larger $K$ is, the more accurate the online model selection) and the principal direction sensing (the larger $K$ is, the smaller $M-K$ principal direction measurements). The MSE of SCS using fully random sensing measurements is plotted in the same figure (blue) for comparison.  The lowest point is attained at $K=9$, where the MSE of the online SCS is $0.65$ times that of SCS. The online SCS thus considerably reduces the MSE of SCS. 

Monte Carlo simulations further show that a similar U-shape graph is obtained with different $M$ values and the Gaussian eigenvalue decay parameter $\alpha$: The online SCS has the lowest MSE with $K$ in the order of $M/2$, and the ratio between the MSE of the online adaptive SCS and the standard SCS is smaller as $\alpha$ and $M$ increase. 

\vspace{-3ex}
\section{Experiments with Real Images}
\label{sec:experiments}
\vspace{-1ex}

The online adaptive SCS is applied in real image sensing, and compared with SCS using fully random measurements. The latter has been reported to bring about 0.5 to 3.5 dB improvement in PSNR at various sample rates with respect to conventional CS based on sparse models~\cite{yu2011SCS}.

\begin{figure}[htbp]
\vspace{1ex}
\begin{center}
\begin{tabular}{c}
\includegraphics[width=6cm]{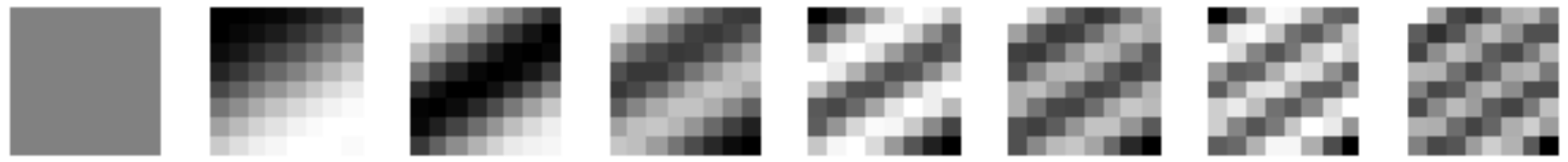}  
\end{tabular}
\end{center}
\vspace{-4ex}
\caption{\small{The first eight principal direction vectors of a directional PCA.}} 
\label{fig:directional:PCA}
\vspace{-1ex}
\end{figure}

\begin{figure}[htbp]
\vspace{0ex}
\begin{center}
\begin{tabular}{ccc}
\hspace{-1ex}
\includegraphics[width=2.5cm]{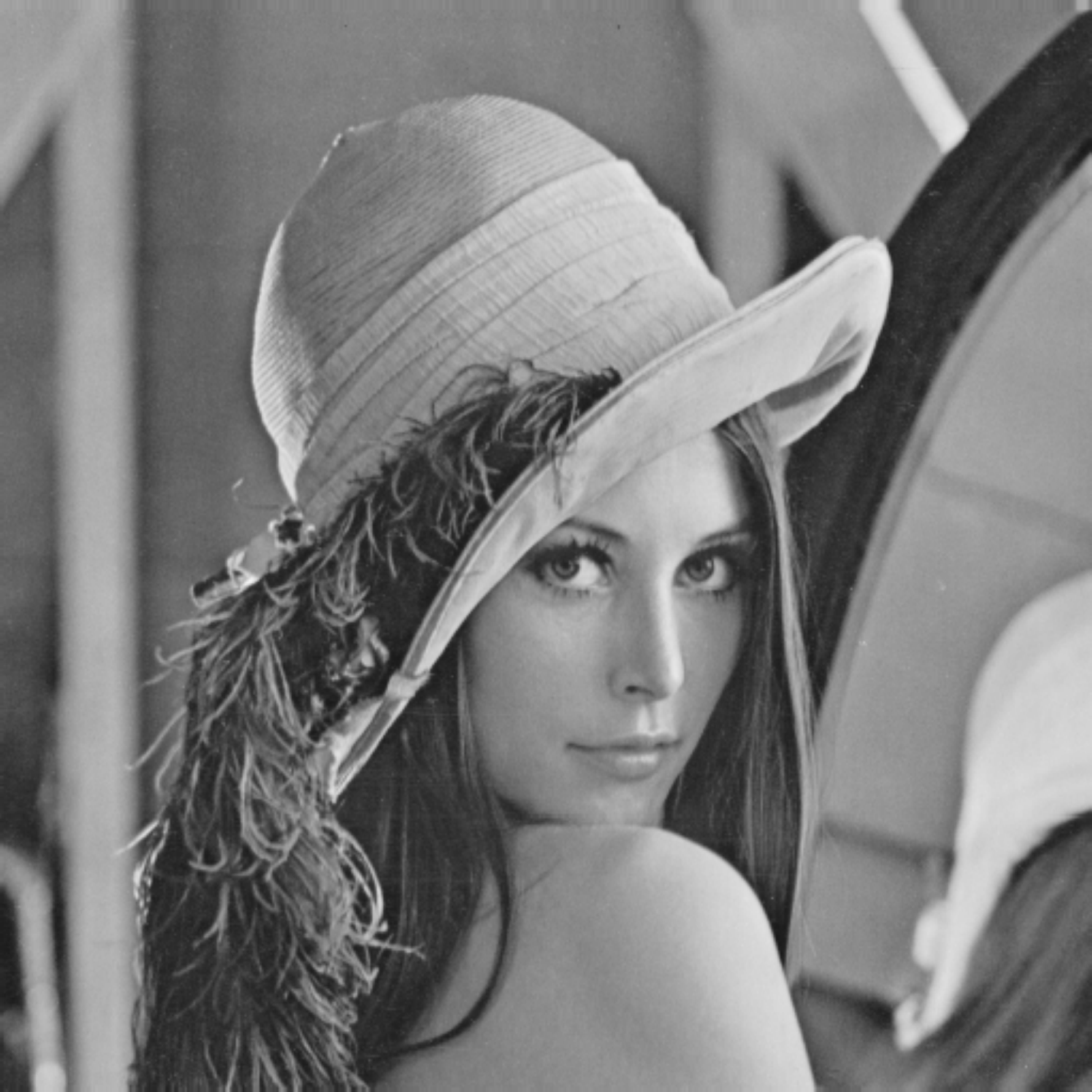}  
\includegraphics[width=2.5cm]{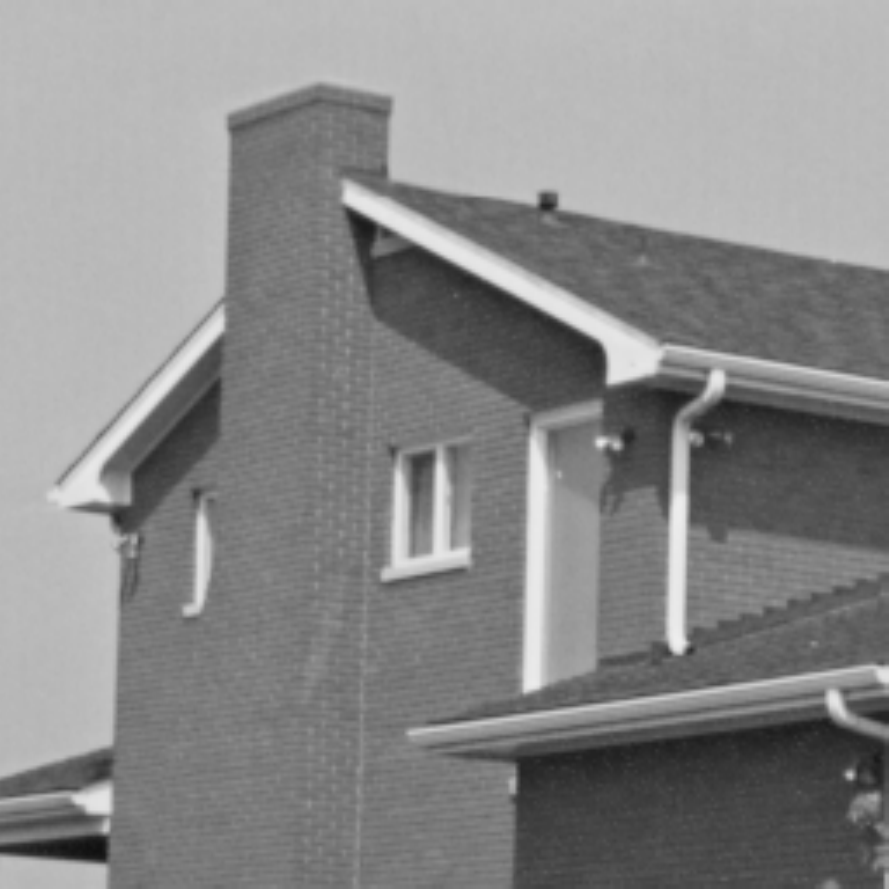}  
\includegraphics[width=2.5cm]{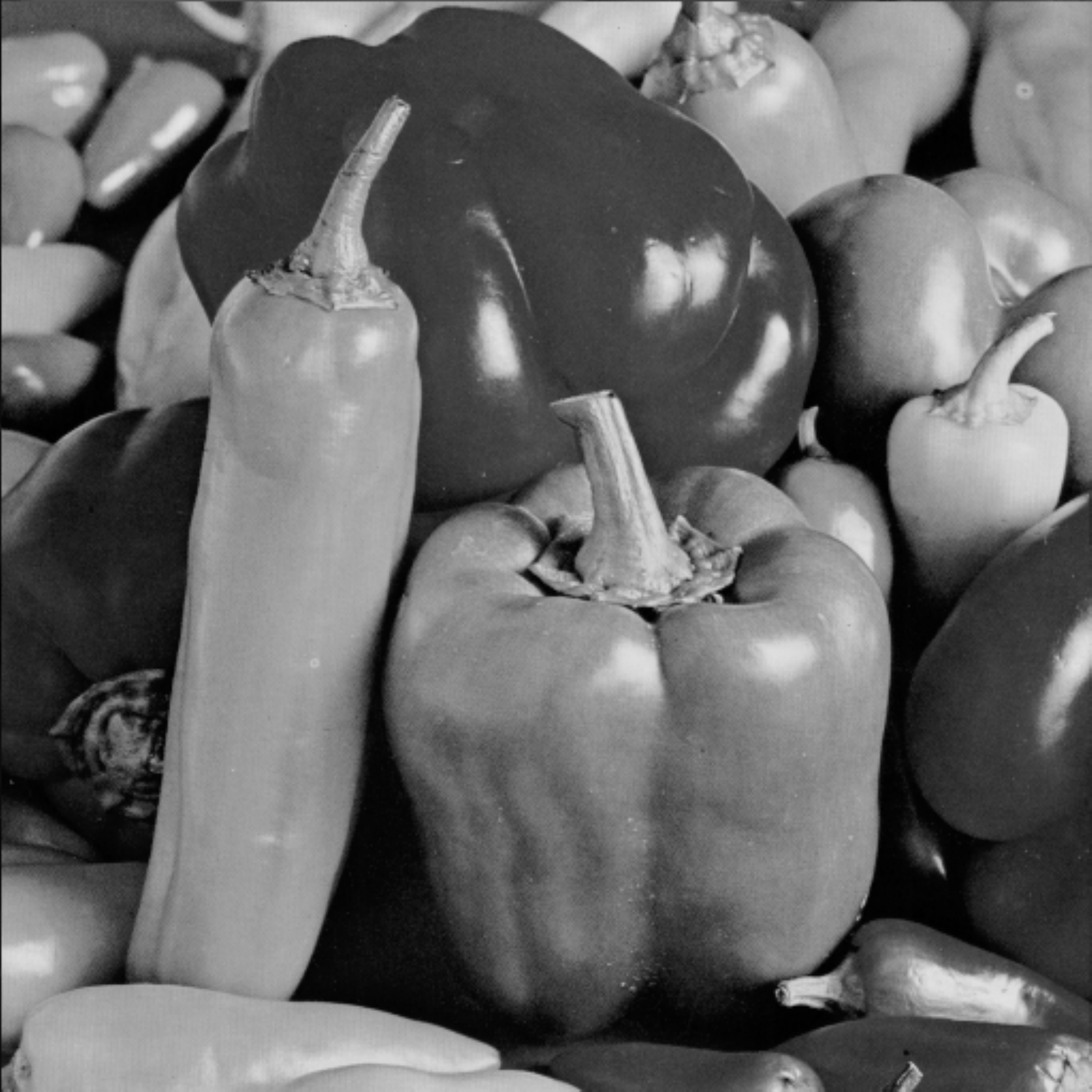}  \\
\end{tabular}
\end{center}
\vspace{-4ex}
\caption{\small{From left to right: Lena, House, and Peppers.}} \label{fig:standard:images} 
\vspace{-3ex}
\end{figure}

Following a common practice, an image is decomposed into $\sqrt{N} \times \sqrt{N} = 8 \times 8$ \textit{non-overlapping} local patches (an image patch is reshaped and considered as a vector), each regarded as a signal and assumed to follow a GMM~\cite{yu2010PLE}. As illustrated in Figure~\ref{fig:directional:PCA}, the GMM is comprised of $J=19$ geometry-motivated Gaussian models, each capturing a local direction (see~\cite{yu2010PLE} for more details). $M=16$ measurements, or equivalent a sampling rate of $M/N=1/4$, are applied. The standard images Lena ($512 \times 512$), House ($256\times 256$), and Peppers ($512 \times 512$), as shown in Figure~\ref{fig:standard:images}, are used in the experiments. 

\begin{figure}[htbp]
\vspace{-2ex}
\begin{center}
\begin{tabular}{c}
\hspace{-5ex}
\includegraphics[width=10cm]{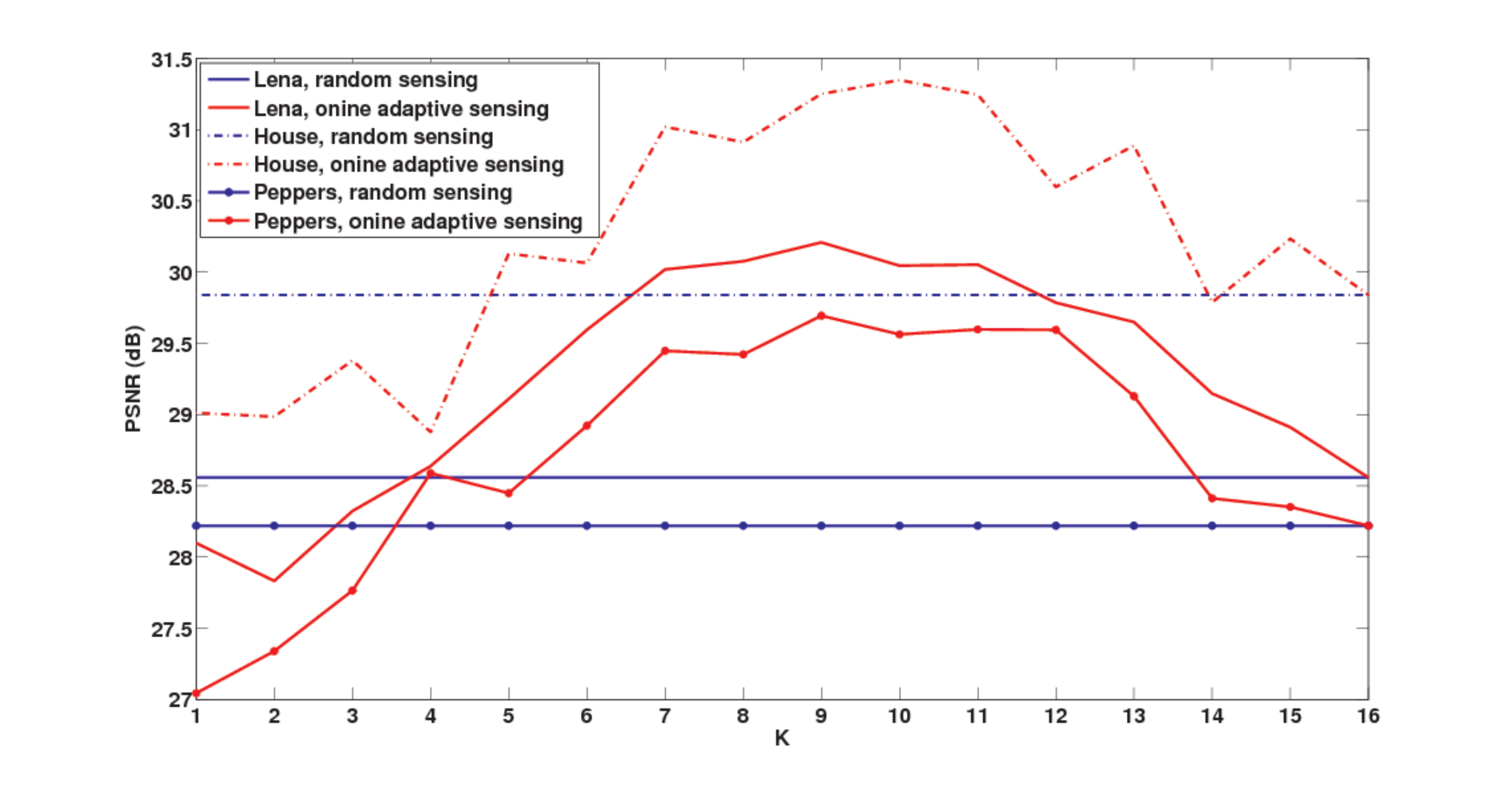}  
\end{tabular}
\end{center}
\vspace{-4ex}
\caption{\small{PSNR comparison of the proposed online adaptive SCS and standard SCS as a function $K$, for three images under test.}} 
\label{fig:PSNR:onlineSCS:vs:SCS}
\vspace{-1ex}
\end{figure}

Figure~\ref{fig:PSNR:onlineSCS:vs:SCS} plots the PSNR of the reconstructed patches obtained with the proposed online adaptive SCS as a function of $K$, the number of first-step random measurements, in comparison with that of standard SCS. Similar to the U-shape curve obtained on the synthetic data in Section~\ref{subsec:performance:analysis}, for all the three images under test, as $K$ increases the PSNR of the adaptive SCS overall first rises, and then decreases, converging to that of the standard SCS as $K$ goes to $M$. The largest improvement with respect to standard SCS, about $1.5$ dB, is attained at $K=9$ or $10$.


\vspace{-2ex}
\section{Conclusion and Future Works}
\label{sec:conclusion}
\vspace{-1ex}

An online adaptive sensing strategy has been developed for signals following a mixture model. The basic idea is to first detect online the model, and then adapt the sensing for it. 
Illustrated for GMMs, the framework considerably reduces the average reconstruction error with respect to standard CS using fully random measurements on both synthetic and real image data, at marginally increased complexity. 

We are currently refining the proposed algorithm. The hard switch from random sensing to optimal sensing triggered by the online model selection may be improved with a sample-per-sample optimization following~\eqref{eqn:online:SCS:MSE}, or extending the analysis developed in~\cite{Carson2011}. 

The proposed scheme imbeds low-level pattern recognition (model selection) in the signal sensing and estimation problem. The pattern recognition part has value by itself, and will be further explored.

Following the recent results in~\cite{Carson2011}, the same type of adaptive sensing strategy can be applied to mixtures of other distributions.\\

\noindent
{\it \small{{\bf Acknowledgments:} Work supported by NSF, ONR, NGA, ARO, DARPA, and NSSEFF. We thank Prof. Robert Calderbank for discussion on the topics of this paper.}}

\vspace{-3ex}

{\small
\bibliographystyle{plain}
\bibliography{biblio}}

\end{document}